\documentclass[journal]{IEEEtran}

\usepackage{cite}
\usepackage{amsmath,amssymb,amsfonts}
\usepackage{graphicx}
\usepackage{textcomp}
\usepackage{xcolor}
\usepackage{booktabs}
\usepackage{array}
\usepackage{url}

\usepackage{xurl}
\newcommand{\hecras}{HEC\mbox{-}RAS}
\newcommand{\wse}{WSE}
\newcommand{\rmse}{RMSE}

\begin{document}

% ------------------------------------------------------------
% Title and authors
% ------------------------------------------------------------
\title{Learned Response-Field Inertia Operator for HEC-RAS 2D Water-Surface Elevation Prediction}

\author{Edward Holmberg, Elias Ioup, Md Meftahul Ferdaus, Mahdi Abdelguerfi, and Julian Simeonov%
\thanks{Manuscript received Month Day, 2026; revised Month Day, 2026.}%
\thanks{Edward Holmberg, Md Meftahul Ferdaus, and Mahdi Abdelguerfi are with the Canizaro Livingston Gulf States Center for Environmental Informatics, Department of Computer Science, The University of New Orleans, New Orleans, LA 70148, USA (e-mail: eholmber@uno.edu; mferdaus@uno.edu; gulfsceidirector@uno.edu).}%
\thanks{Elias Ioup is with the Center for Geospatial Sciences, Naval Research Laboratory, Mississippi, USA (e-mail: elias.z.ioup.civ@us.navy.mil).}%
\thanks{Julian Simeonov is with the Ocean Sciences Division, Naval Research Laboratory, Mississippi, USA (e-mail: julian.a.simeonov.civ@us.navy.mil).}%
}

\markboth{IEEE Transactions on [Journal Name],~Vol.~XX, No.~X, Month~2026}%
{Holmberg \MakeLowercase{\textit{et al.}}: LRFIO for HEC-RAS 2D Water-Surface Elevation Prediction}

\maketitle

% ------------------------------------------------------------
% Abstract and keywords
% ------------------------------------------------------------

\begin{abstract}
This article presents a cross-dataset evaluation of learned native-cell surrogate models for solver-consistent water-surface elevation (\wse{}) prediction in \hecras{} 2D. To avoid raster remapping error and information-access confounding, surrogates are evaluated directly on the original nonuniform computational cells under an explicit policy that separates static project inputs, current hydraulic state, project-input forcing, calibration-derived quantities, and future solver-output targets. We introduce the Learned Response-Field Inertia Operator (LRFIO), a no-forcing, increment-based learned surrogate that calibrates an inertial response operator from solved \hecras{} trajectories and deploys the retained operator through closed-form native-cell rollout. LRFIO evaluates a base-case-first response hierarchy consisting of persistence, global calibrated inertia, and segmented response-field inertia. Segmentation, residual correction, and neuralized inertia are treated as learnable modeling choices, with added complexity retained only when validation evidence justifies its cost. Evaluated across four diverse \hecras{} 2D benchmarks, LRFIO retains different response structures for different domains, demonstrating adaptive learned complexity. The selector audit shows controlled complexity with a maximum validation regret of 4.30\%. During deployment, retained rollout times range from 0.003~s to 0.242~s, and the Beaver Bayou measured-solve comparison gives an estimated $(2.75\times10^{4})$ horizon-normalized speedup over \hecras{}. These results indicate that the current native-cell increment is a strong solver-conditioned predictive scaffold and that added response-field, neural, or spatial complexity should be retained only when empirically justified.
\end{abstract}


% ============================================================

\section{Introduction}
\label{sec:introduction}

\IEEEPARstart{T}{wo-dimensional} hydrodynamic models such as \hecras{} 2D are widely used to simulate flood inundation and hydraulic response over complex terrain~\cite{USACE_HECRAS_2DUsersManual,USACE_HECRAS_NumericalMethods}. Because repeated numerical solves are computationally expensive, surrogate modeling has become an established strategy for optimization, uncertainty analysis, scenario screening, and real-time or near-real-time workflows~\cite{Razavi2012_SurrogateReview}. This article studies learned surrogate modeling for \hecras{} 2D unsteady-flow simulations in a solver-consistent setting. The objective is solver-consistent emulation of future \hecras{} water-surface elevation (\wse{}) fields directly on the original nonuniform computational cells, the native cells defined by the project geometry and discretization, rather than on rasterized image grids~\cite{USACE_HECRAS_2DUsersManual,USACE_HECRAS_NumericalMethods}.

Recent scientific machine learning approaches use raster convolutions, recurrent sequence models, graph neural networks, and neural operators to emulate flood states~\cite{Li2021_FNO,Lu2021_DeepONet,Bentivoglio2023_SWEGNN,Bentivoglio2025_mSWEGNN}. These approaches are valuable, but direct comparisons across surrogate models can be misleading when spatial representation, forecast protocol, and information access differ. Projecting irregular native cells to raster grids can introduce support mismatch and remapping error, while giving a model full-horizon future project-input forcing changes the deployment problem relative to current-state-only prediction~\cite{USACE_HECRAS_BoundaryConditions}. This study therefore treats surrogate modeling as a controlled component-evaluation problem in which representation, information access, and model capacity are recorded explicitly rather than collapsed into an undifferentiated model leaderboard.

The motivating machine-learning question is how much learned model capacity is required once the native-cell hydraulic state is represented in a solver-consistent form. A central observation is that the current native-cell \wse{} increment,

\begin{equation}
\Delta \mathbf{W}_t
=
\mathbf{W}_t-\mathbf{W}_{t-1}.
\label{eq:intro_current_increment}
\end{equation}

summarizes the recent response of the solved \hecras{} system after routing water through terrain, storage, connectivity, boundary conditions, and numerical controls~\cite{USACE_HECRAS_NumericalMethods}. This makes the increment a compact learned-surrogate scaffold: the surrogate can use the most recent solver response directly and learn how that response persists, decays, saturates, or varies across the native-cell domain.

This motivates framing future \wse{} prediction as an increment-response learning problem. For initialization time $t$, increment-based surrogates reconstruct future \wse{} over a horizon $H$ as

\begin{equation}
\widehat{\mathbf{W}}_{t+k}
=
\mathbf{W}_t
+
\sum_{j=1}^{k}
\widehat{\Delta \mathbf{W}}_{t+j},
\quad
k=1,\ldots,H.
\label{eq:intro_increment_reconstruction}
\end{equation}

This inertial-continuation view is conceptually related to simplified local-inertial approximations of flood-wave propagation~\cite{Bates2010_SimpleInertial,Almeida2013_LocalInertialApplicability}. The method developed here uses this idea as a surrogate-learning scaffold: it learns an effective response operator from solved \hecras{} trajectories and evaluates, through ablation and validation-based selection, how much model complexity is required for accurate native-cell rollout.

To exploit this increment scaffold, this article introduces the \emph{Learned Response-Field Inertia Operator} (LRFIO), a no-forcing native-cell surrogate that learns an inertial response operator from solved \hecras{} trajectories. LRFIO evaluates a response hierarchy consisting of persistence, global calibrated inertia, and segmented response-field inertia, then retains the simplest response structure supported by validation evidence. Segmentation, residual correction, and neuralized inertia are treated as learnable modeling choices that are retained only when validation evidence justifies their added cost.

This framing is central to the machine-learning interpretation of the method. LRFIO learns response partitions, inertial coefficients, increment caps, and selector decisions from allowed training and validation data. The retained operator is then deployed through closed-form inference using only the current and previous native-cell \wse{} fields. LRFIO contributes an ablation-driven learned response operator that compresses solver-derived predictive structure into a compact, validation-justified deployment form.

LRFIO is evaluated across four \hecras{} 2D datasets: Beaver Bayou~\cite{USACE_NOLA_BeaverBayou_PrivateDataset}, Upper San Saba River~\cite{Dryad_UpperSanSaba_Harris2023}, Lower San Saba River~\cite{Dryad_LowerSanSaba_Harris2024}, and Tuttle Creek / Big Blue / Kansas River~\cite{Dryad_TuttleCreek_Wiest2024}. All experiments use chronological splits and a leakage-avoidance protocol that separates static inputs, current state, project-input forcing, calibration-derived quantities, and future solver-output targets. This protocol allows no-forcing current-state surrogates, forcing-aware variants, neural-operator baselines, graph models, raster models, and inertial references to be interpreted according to both architecture and information access.

This article addresses three gaps: learned \hecras{} surrogates need native-cell evaluation, architecture comparisons must separate information access from model capacity, and neural or spatial complexity should be justified against strong learned inertial baselines. Accordingly, this article makes the following contributions:
\begin{enumerate}
\item Establishes a leakage-aware native-cell \hecras{} 2D surrogate-learning protocol across four datasets, strictly separating static inputs, current states, project-input forcing, calibration-derived quantities, and future solver-output targets.
\item Formulates \wse{} prediction directly on native \hecras{} computational cells, avoiding raster remapping error and aligning surrogate evaluation with the reference solver support.
\item Introduces LRFIO as an ablation-derived learned response operator that calibrates inertial response structure from solved \hecras{} trajectories and deploys the retained operator through closed-form native-cell inference.
\item Evaluates a broad component space spanning persistence, global and segmented calibrated inertia, neuralized inertia variants, graph and Fourier models, raster networks, and forcing-conditioned competitors.
\item Demonstrates through cross-dataset selector audits and speed--accuracy comparisons that the current native-cell increment is a strong solver-conditioned predictive scaffold, including a Beaver Bayou horizon-normalized speedup of $2.75 \times 10^4$ over the measured \hecras{} solve while retaining response-field complexity only when empirically justified.
\end{enumerate}

The remainder of the article is organized as follows. Section~\ref{sec:background_related_work} reviews related work. Section~\ref{sec:problem_data_protocol} defines the forecasting problem, information-access policy, and benchmark datasets. Section~\ref{sec:irfio} details LRFIO. Section~\ref{sec:experimental_methodology} describes the evaluation protocol. Section~\ref{sec:results} reports cross-dataset results. Section~\ref{sec:discussion} discusses implications and future work, and Section~\ref{sec:conclusion} concludes.
% ============================================================

% =================================== =========================

\section{Background and Related Work}
\label{sec:background_related_work}

\subsection{HEC-RAS Emulation and Surrogate Representations}
\label{subsec:background_hecras_representations}

\hecras{} 2D is widely used to simulate unsteady hydraulic responses over project-defined, nonuniform computational cells~\cite{USACE_HECRAS_2DUsersManual,USACE_HECRAS_NumericalMethods,USACE_HECRAS_2DOptions}. In surrogate modeling, the numerical solver is typically treated as the reference system to be emulated, which requires a strict separation between static project inputs, solver-output targets, and boundary-condition forcing~\cite{USACE_HECRAS_WorkingWithProjects,USACE_HECRAS_BoundaryConditions}. While surrogate approaches are well established for reducing the computational burden of repeated hydraulic simulations~\cite{Razavi2012_SurrogateReview}, their reported performance depends on model architecture, spatial representation, forecast protocol, and the information available at prediction time.

Recent data-driven flood surrogates employ architectures ranging from gradient-boosted trees to deep convolutional, recurrent, graph-based, and neural-operator models~\cite{HacesGarcia2023_DLHFRAN,Liao2023_UrbanPluvialFlood_CNN,Xu2023_LightGBM_UrbanFlood,Karapetyan2026_UNet_CNNLSTM_ConvLSTM,Wang2026_Hybrid_CNN_LSTM_CoastalFlood}. Rasterizing irregular \hecras{} cells into grid tensors enables standard CNN, U-Net, and ConvLSTM applications~\cite{Ronneberger2015_UNet,Shi2015_ConvLSTM}, while also introducing many-to-one aggregation, empty-grid support, and native-to-raster remapping effects that must be measured when interpreting model accuracy. Mesh-based graph neural networks address irregular support by learning message-passing updates directly on nodes or cells~\cite{Pfaff2021_MeshGraphNets,Bentivoglio2023_SWEGNN,Bentivoglio2025_mSWEGNN}. Neural operators such as Fourier Neural Operators and DeepONet learn mappings in spectral or continuous function spaces~\cite{Li2021_FNO,Lu2021_DeepONet}, with recent variants incorporating non-rectangular geometries~\cite{Li2023_GeoFNO,Li2023_GINO}.

These models demonstrate the value of machine learning for hydrodynamic emulation and motivate component-level evaluation of where predictive skill originates. In native-cell \hecras{} emulation, architectural gains are easiest to interpret when spatial support, information access, and model complexity are made explicit. The present study therefore evaluates surrogate models as controlled components of a solver-emulation workflow, with separate labels for native-cell support, raster or graph representation, current-state versus forcing-aware prediction, and learned response complexity.

\subsection{Reduced-Order, Sequence, and Learned Inertial Surrogates}
\label{subsec:background_reduced_order_inertia}

Complementing deep spatial networks, reduced-order modeling provides efficient approximations by representing high-dimensional hydraulic fields in lower-dimensional subspaces using proper orthogonal, dynamic mode, or non-intrusive decompositions~\cite{Schmid2010_DMD,Bistrian2015_DMDPODShallowWater,Dutta2021_PODRBF_NIROM}. State-space and sequence models, including LSTMs and GRUs, predict future behavior by tracking recent residuals, increments, or latent state changes over time~\cite{Hochreiter1997_LSTM,Cho2014_GRU}. These methods are relevant because \wse{} prediction is inherently temporal: the forecast depends on the present state and on how the solved hydraulic system has recently changed.

From a hydrodynamic perspective, simple-inertial and local-inertial formulations efficiently approximate shallow-water behavior by retaining a simplified representation of momentum persistence~\cite{Bates2010_SimpleInertial,Almeida2013_LocalInertialApplicability,Sridharan2021_LocalInertialUnstructured,NithilaDevi2024_SubgridLocalInertial}. The present study uses a related surrogate-learning principle: the recent solver-produced \wse{} increment can be treated as an effective response signal generated by the full \hecras{} model after terrain, storage, roughness, boundary conditions, and numerical controls have influenced the solution.

This distinction frames the proposed method as a learned inertial surrogate. Solved trajectories provide evidence for estimating how the current native-cell increment should persist, decay, saturate, or vary across response regions during rollout. The learned quantities may include response partitions, inertial coefficients, increment caps, and validation-based selector decisions. Once these quantities are learned, deployment can be closed form because the learned operator has been compressed into a compact response rule whose parameters and structure were selected from data.

\subsection{Ablation, Complexity Control, and Information Access}
\label{subsec:background_ablation_complexity}

A recurring challenge in scientific machine learning is distinguishing useful model capacity from unnecessary complexity. High-capacity networks can fit complex spatial and temporal responses, yet added layers, latent states, or learned corrections must improve held-out deployment behavior to justify their cost. Controlled ablation provides the mechanism for this decision: neural correction layers, graph encoders, raster representations, segmented response fields, and simpler learned operators can be compared under the same information-access policy, allowing the retained surrogate to be selected from validation evidence.

This issue is especially important for flood and river-model surrogates because information access can change the task itself. Models conditioned on full-horizon future forcing~\cite{HacesGarcia2023_DLHFRAN,Wang2026_Hybrid_CNN_LSTM_CoastalFlood} answer a different deployment question than no-forcing current-state models. Boundary hydrographs, rainfall series, lateral inflows, and stage hydrographs are legitimate project inputs~\cite{USACE_HECRAS_BoundaryConditions}, and explicit forcing-access labels are needed so that improvements can be attributed to architecture, input information, or both.

The proposed evaluation protocol therefore separates three questions that are often conflated. First, what spatial support is used: raster grids, graph nodes, modal bases, or native \hecras{} cells? Second, what information is available at forecast initialization: current state only, step-ahead forcing, or full-horizon forcing? Third, how much learned model complexity is justified: persistence, global calibrated inertia, segmented response-field inertia, or neural/operator correction? This separation positions LRFIO as a learned surrogate operator selected by validation evidence under an explicit information-access policy.

\subsection{Position of the Present Study}
\label{subsec:background_position}

Despite the proliferation of high-capacity surrogates~\cite{Razavi2012_SurrogateReview,Liao2023_UrbanPluvialFlood_CNN,Bentivoglio2023_SWEGNN,Bentivoglio2025_mSWEGNN,Li2021_FNO,Lu2021_DeepONet,Schmid2010_DMD,Bates2010_SimpleInertial}, several gaps remain for native-cell \hecras{} 2D prediction. Evaluations often conflate architecture improvements with information access, raster or grid-based surrogates can introduce representation effects before learning begins, and complex neural architectures are rarely benchmarked directly against strong learned inertial baselines across multiple \hecras{} datasets. These gaps make it difficult to determine when deep spatial or neural correction layers are necessary for solver-consistent native-cell rollout.

This work is also connected to a broader research lineage in geospatial data management, river-channel geometry representation, spatial indexing, and spatiotemporal information systems~\cite{Abdelguerfi2001_3DSyntheticEnvironment,Wilson2003_GIDB_XML,Flanagin2007_HydraulicSplines,Ioup2007_AKNN_MTree,Ladner2012_SpatioTemporalMining}. More recent related work has explored recurrent neural-operator acceleration for \hecras{} river forecasting~\cite{Holmberg2025_HECRAS_RNO}. The present article builds on that surrogate-modeling direction by asking a sharper ablation question: when the current native-cell increment is preserved as a solver-conditioned response signal, how much additional neural or spatial complexity improves the retained deployment tradeoff?

This article addresses the identified gaps by introducing the Learned Response-Field Inertia Operator (LRFIO), a native-cell surrogate that learns a compact inertial response operator from solved \hecras{} trajectories. LRFIO evaluates a hierarchy of response hypotheses, including persistence, global calibrated inertia, and segmented response-field inertia, and retains the simplest sufficient structure under a leakage-controlled validation protocol. By isolating native-cell performance and explicitly tracking information access, this study tests calibrated learned inertia as a strong baseline before attributing predictive value to higher-capacity neural, graph, raster, or operator architectures.
% ============================================================

% ============================================================
\begin{figure*}[!t]
\centering
\includegraphics[width=\textwidth]{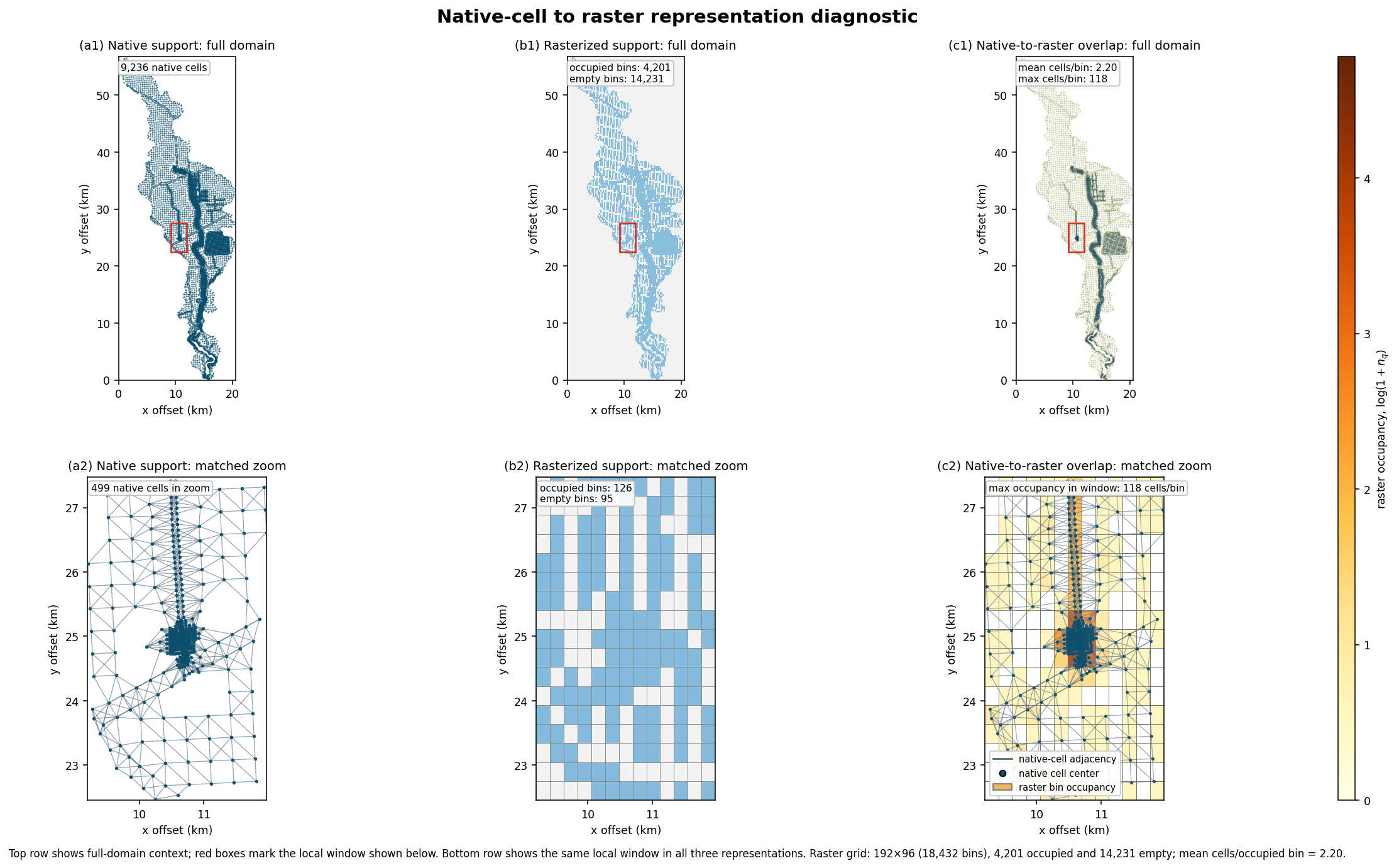}
\caption{Representative native-cell to raster representation diagnostic for Beaver Bayou.
The top row shows full-domain context, and the red boxes mark the local window magnified in the bottom row.
(a1--a2) Native \hecras{} cell support illustrates the irregular computational geometry.
(b1--b2) Rasterization projects the same geometry onto a regular grid, producing occupied and empty bins.
(c1--c2) The overlap view shows how multiple native cells map into the same raster bin, producing many-to-one aggregation.
The diagnostic motivates direct native-cell evaluation and serves as a representation analysis rather than a surrogate-model result.}
\label{fig:native_vs_raster_adapter}
\end{figure*}

\section{Problem Formulation and Benchmark Data Protocol}
\label{sec:problem_data_protocol}

\subsection{Forecasting Task and Native-Cell Surrogate Representation}
\label{subsec:problem_reference_forecasting_task}
\label{subsec:problem_data_canonical_arrays}

Let $d$ index an evaluated \hecras{} 2D unsteady-flow case. Each case defines a numerical reference mapping from project-defined scenario information to time-indexed hydraulic result fields over a two-dimensional flow area~\cite{USACE_HECRAS_2DUsersManual,USACE_HECRAS_NumericalMethods,USACE_RASMapper_ResultsMaps}. The surrogate-learning target is solver-consistent emulation of future \hecras{} water-surface elevation (\wse{}) fields under a declared input policy. The task is supervised reference-solver emulation: models learn to reproduce future \hecras{} output fields from information available at forecast initialization while preserving the native computational-cell support of the original project.

For dataset $d$, let $N_d$ be the number of native \hecras{} 2D computational cells and $T_d$ the number of available output frames. The canonical native-cell \wse{} matrix is
\begin{equation}
\mathbf{W}^{(d)}
\in
\mathbb{R}^{T_d \times N_d},
\label{eq:data_wse_matrix}
\end{equation}
where each row is an output time and each column is a native computational cell. Associated static geometry arrays include cell-center coordinates
$\mathbf{X}^{(d)}\in\mathbb{R}^{N_d\times2}$ and bed elevations
$\mathbf{z}^{(d)}\in\mathbb{R}^{N_d}$. When available, additional static project features such as roughness, cell area, connectivity, and boundary geometry are also extracted.

Depth and water-surface increments are derived consistently on the native cells:
\begin{align}
d^{(d)}_{t,i}
&=
\max\left(W^{(d)}_{t,i}-z^{(d)}_i,0\right),
\label{eq:data_depth_definition}\\
\Delta \mathbf{W}^{(d)}_t
&=
\mathbf{W}^{(d)}_t-\mathbf{W}^{(d)}_{t-1}.
\label{eq:problem_current_increment}
\end{align}
The increment field is retained as a canonical dynamic descriptor because it records recent solver-produced hydraulic change on the native computational support. In the learned-response framing used by this article, $\Delta \mathbf{W}^{(d)}_t$ is the observable state signal from which candidate surrogate operators learn how the recent solver response persists, decays, saturates, or varies across cells during future rollout.

For an initialization time $t_0$ and forecast horizon $H$, the surrogate predicts the future native-cell sequence
\begin{equation}
\widehat{\mathbf{W}}_{t_0+1:t_0+H}
=
\left\{
\widehat{\mathbf{W}}_{t_0+1},
\ldots,
\widehat{\mathbf{W}}_{t_0+H}
\right\}.
\label{eq:problem_forecast_sequence}
\end{equation}
A general surrogate family is written as
\begin{equation}
\widehat{\mathbf{W}}_{t_0+1:t_0+H}
=
\mathcal{M}_{\Theta}\left(\mathcal{I}_{t_0}\right),
\label{eq:problem_general_surrogate}
\end{equation}
where $\mathcal{M}_{\Theta}$ is a calibrated or learned model, $\Theta$ denotes learned or calibration-derived quantities, and $\mathcal{I}_{t_0}$ is the model's declared input set at forecast initialization. This notation is intentionally broad: $\mathcal{M}_{\Theta}$ may represent a neural network, graph model, raster model, reduced-order model, learned inertial operator, or closed-form response rule whose parameters and structure were estimated from data.

Increment-based models reconstruct future \wse{} by accumulating predicted increments:
\begin{equation}
\widehat{\mathbf{W}}_{t_0+k}
=
\mathbf{W}_{t_0}
+
\sum_{j=1}^{k}
\widehat{\Delta\mathbf{W}}_{t_0+j},
\quad k=1,\ldots,H.
\label{eq:problem_increment_reconstruction}
\end{equation}

Candidate rows may differ in representation, forecast protocol, update form, spatial or response structure, neural correction, and forcing access. These distinctions are recorded as model metadata so that accuracy and runtime are interpreted in relation to the task each model is allowed to solve. A high-capacity neural model, a forcing-aware sequence model, and a no-forcing learned inertial operator may all report \wse{} error on the same held-out horizon, but they can differ substantially in spatial support, available information, and deployment assumptions.

\subsection{Information-Access Policy}
\label{subsec:problem_information_access}

To prevent future-target leakage and separate architecture effects from scenario-information access, model inputs are partitioned into allowable and disallowed categories~\cite{Kaufman2012_LeakageDataMining}. This partition is central to the surrogate-learning interpretation of the study because it specifies the information available to each model before architectural comparisons are made.

Allowable information may include:
\begin{enumerate}
\item \textbf{Static project inputs} $\mathcal{S}^{(d)}$: geometry, bed or terrain elevation, roughness or land-cover attributes, cell area, connectivity, boundary geometry, and related project-defined quantities.

\item \textbf{Dynamic initialization state} $\mathcal{X}_{t_0}$: current or past hydraulic state available at forecast initialization, including $\mathbf{W}_{t_0}$, $\mathbf{W}_{t_0-1}$, $\Delta\mathbf{W}_{t_0}$, current depth, wetness, and surrogate-generated states during rollout.

\item \textbf{Project-input forcing} $\mathcal{F}_{t_0+1:t_0+H}$: prescribed scenario-definition information such as boundary-condition hydrographs, stage hydrographs, rainfall, lateral inflow, or unsteady-flow input series. These inputs are available only to rows explicitly labeled as forcing-aware~\cite{USACE_HECRAS_BoundaryConditions}.

\item \textbf{Calibration-derived or learned quantities} $\Theta$: learned weights, response-field maps, response descriptors, scalers, selected hyperparameters, inertial coefficients, increment caps, selector decisions, neural-layer parameters, graph structures, modal bases, or other quantities computed from allowed training or validation data before held-out test evaluation.
\end{enumerate}

Thus, an evaluated model has an input set
\begin{equation}
\mathcal{I}_{t_0}
\subseteq
\left\{
\mathcal{X}_{t_0},
\mathcal{S}^{(d)},
\mathcal{F}_{t_0+1:t_0+H},
\Theta
\right\}.
\label{eq:problem_allowed_inputs}
\end{equation}
No-forcing, step-ahead-forcing, and full-horizon-forcing rows are therefore distinct information-access conditions. Likewise, a closed-form learned operator and a neural network are directly comparable as architecture alternatives only when they use the same initialization information and are selected under the same validation protocol.

The disallowed set includes future \hecras{} result fields and any quantities derived from them:
\begin{equation}
\mathcal{I}_{t_0}
\cap
\left\{
\mathbf{W}_{t_0+k},
\mathbf{d}_{t_0+k},
\mathbf{v}_{t_0+k},
\Delta \mathbf{W}_{t_0+k},
\ldots
\right\}
=
\varnothing,
\quad k\geq 1.
\label{eq:problem_no_future_results}
\end{equation}
Future \wse{}, depth, velocity, wet/dry state, and solver-output-derived deltas are prediction targets or diagnostics only.

The retained Learned Response-Field Inertia Operator (LRFIO) is evaluated as a no-forcing current-state surrogate with deployment input set
\begin{equation}
\mathcal{I}^{\mathrm{LRFIO}}_{t_0}
=
\left\{
\mathbf{W}_{t_0},
\mathbf{W}_{t_0-1},
\Delta\mathbf{W}_{t_0},
\Theta_{\mathrm{LRFIO}}
\right\}.
\label{eq:problem_lrfio_input_set}
\end{equation}
Here, $\Theta_{\mathrm{LRFIO}}$ contains the learned response structure retained before test evaluation, including the selected response case, response-field partition when applicable, inertial coefficients, increment caps, and selector-audit metadata. LRFIO therefore performs closed-form inference at deployment using a response rule learned from training and validation trajectories.

\subsection{Benchmark Datasets}
\label{subsec:problem_data_benchmark_scope}

The benchmark contains four solved \hecras{} 2D datasets. Beaver Bayou was privately provided by the U.S. Army Corps of Engineers, New Orleans District~\cite{USACE_NOLA_BeaverBayou_PrivateDataset}. The remaining three are public, field-calibrated Dryad datasets: Upper San Saba River~\cite{Dryad_UpperSanSaba_Harris2023}, Lower San Saba River~\cite{Dryad_LowerSanSaba_Harris2024}, and Tuttle Creek / Big Blue / Kansas River~\cite{Dryad_TuttleCreek_Wiest2024}. The same canonical loading, chronological splitting, information-access labeling, and metric-reporting protocol is applied to all four datasets.

Table~\ref{tab:dataset_protocol_summary} summarizes the evaluated domains and forecast protocols. The datasets differ substantially in native-cell count, output interval, trajectory length, and physical forecast duration. These differences are useful for surrogate evaluation because they test whether the learned response-operator framework remains coherent across distinct \hecras{} 2D projects rather than a single event, geometry, or grid size.

\begin{table*}[!t]
\caption{Native-cell \hecras{} 2D benchmark datasets and chronological evaluation protocol. Validation and test entries are reported as initialization frame / horizon steps.}
\label{tab:dataset_protocol_summary}
\centering
\renewcommand{\arraystretch}{1.12}
\footnotesize
\setlength{\tabcolsep}{4pt}
\begin{tabular*}{\textwidth}{@{\extracolsep{\fill}}lrrrrccc@{}}
\toprule
\textbf{Dataset} &
\textbf{Cells} &
\textbf{Frames} &
\textbf{$\Delta t$} &
\textbf{Train frames} &
\textbf{Val.} &
\textbf{Test} &
\textbf{Test dur.} \\
\midrule
Beaver Bayou &
9,236 &
119 &
0.5 h &
$[0,78)$ &
77/17 &
94/24 &
12.0 h \\

Upper San Saba River &
430,874 &
289 &
0.25 h &
$[0,193)$ &
192/48 &
240/48 &
12.0 h \\

Lower San Saba River &
201,676 &
577 &
0.25 h &
$[0,481)$ &
480/48 &
528/48 &
12.0 h \\

Tuttle Creek / Big Blue / Kansas River &
44,255 &
121 &
0.1667 h &
$[0,47)$ &
46/37 &
83/37 &
6.1667 h \\
\bottomrule
\end{tabular*}
\end{table*}

\subsection{Native-Cell and Raster Representation Diagnostic}
\label{subsec:rasterization_diagnostic}

Many spatial neural architectures require regular grid tensors, so a deterministic native-cell-to-raster adapter was constructed to quantify the representation burden before model training. Figure~\ref{fig:native_vs_raster_adapter} shows this diagnostic for Beaver Bayou. The adapter maps 9,236 native cells to a $96\times192$ grid with 18,432 raster bins. Only 4,201 bins are occupied, 14,231 are empty, the mean number of native cells per occupied bin is approximately 2.20, and the maximum is 117. A native-cell to raster to native-cell round trip yields a mean stage \rmse{} of approximately 0.961 m and a maximum-frame \rmse{} of approximately 1.015 m before model training. These values show that rasterization can substantially change the computational support before any surrogate model is trained.

The diagnostic is representative and is used to motivate native-cell metric reporting and representation-aware interpretation of raster model rows. In the context of the present study, this diagnostic clarifies the value of evaluating learned response operators directly on the solver support, where predictions and errors remain aligned with the original \hecras{} computational cells.

\subsection{Evaluation Objective}
\label{subsec:problem_evaluation_objective}

Each dataset is partitioned chronologically into training, validation, and held-out test regions. This separation follows standard supervised-learning practice for reducing selection bias in reported performance~\cite{Hastie2009_ESL}. Training data are used to estimate model parameters, response features, graph structures, modal bases, neural weights, or inertial-response quantities. Validation data are used for hyperparameter selection, family-level decisions, ablation decisions, and guardrail checks. Held-out test data are used only for final evaluation.

The objective is to evaluate surrogate families that predict $\widehat{\mathbf{W}}_{t_0+1:t_0+H}$ from allowed information available at or before forecast initialization while operating on native \hecras{} cells and avoiding future solver-output leakage. Model families are compared by native-cell accuracy, final-lead behavior, tail and hotspot behavior, runtime, information-access condition, retained complexity, and cross-dataset consistency.

This protocol supports the central ablation question of the article: after the current native-cell increment is preserved as a solver-conditioned response signal, how much additional learned complexity is justified? The evaluated space therefore includes simple references, learned global and segmented inertial response operators, neuralized inertia variants, and higher-capacity graph, raster, Fourier, recurrent, and forcing-conditioned competitors. Formal metric definitions and the aggregate selection score are given in Section~\ref{sec:experimental_methodology}.
% ============================================================

\section{Learned Response-Field Inertia Operator}
\label{sec:irfio}

\begin{figure*}[!t]
\centering
\includegraphics[width=\textwidth]{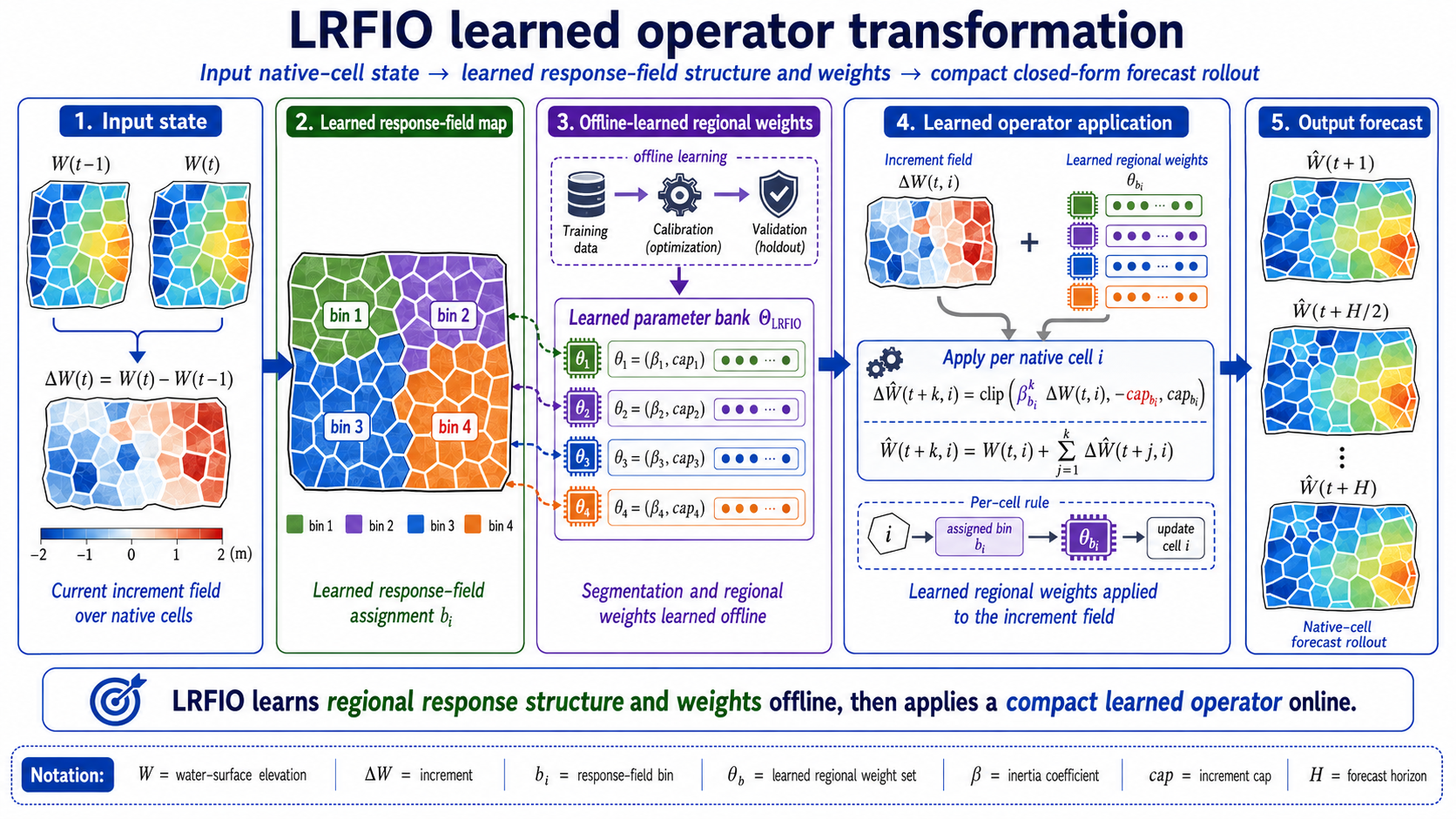}
\caption{Learned Response-Field Inertia Operator (LRFIO) operator transformation.
The figure shows the retained deployment pathway from native-cell state history to closed-form forecast rollout.
The current and previous native-cell water-surface fields form the increment field, cells are assigned to learned response-field bins, offline-learned regional weights $\theta_b=(\beta_b,\mathrm{cap}_b)$ are retrieved, and the retained learned operator applies a capped inertial update over the forecast horizon.
The figure emphasizes the asymmetric structure of LRFIO: response-field structure and weights are learned offline, while online deployment applies the compact retained operator directly on native cells.}
\label{fig:lrfio_operator_transformation}
\end{figure*}

\subsection{Architecture Overview and Learned Response Hierarchy}
\label{subsec:irfio_overview}

The retained architecture is the Learned Response-Field Inertia Operator (LRFIO). Figure~\ref{fig:lrfio_operator_transformation} presents LRFIO as a learned operator transformation from the current native-cell hydraulic state to a closed-form forecast rollout. The input to the operator is the most recent native-cell water-surface history, represented by $W_(t-1)$ and $W_(t)$. Their difference forms the current solver-produced increment field,
\begin{equation}
\Delta \mathbf{W}_t = \mathbf{W}_t - \mathbf{W}_{t-1},
\label{eq:lrfio_current_increment}
\end{equation}
which acts as the dynamic state signal propagated by the retained response operator.

The key modeling premise is that $\Delta W(t)$ is used as a solver-conditioned response signal. In the solver-consistent setting, this increment is the most recent native-cell response produced by \hecras{} after numerical routing through terrain, storage, roughness, boundary conditions, and project-specific geometry. LRFIO directs learning toward how this response signal should persist, decay, saturate, or vary across the native-cell domain during future rollout.

LRFIO learns two coupled objects from solved \hecras{} trajectories before held-out testing. First, when segmented response structure is retained, it learns a response-field assignment $b_i$ that maps each native cell $i$ to a response region. Second, it learns a regional weight set
\begin{equation}
\theta_b=(\beta_b,\mathrm{cap}_b),
\label{eq:lrfio_regional_weight_set}
\end{equation}
where $\beta_b$ controls how the current increment persists or decays across forecast lead time and $\mathrm{cap}_b$ constrains the magnitude of the propagated increment. The retained learned object is a compact regional response operator whose structure, cell assignments, and parameters are estimated from training and validation evidence.

During online deployment, the retained operator applies the learned regional weights to each native cell according to its response-field assignment. For a lead $k$ and native cell $i$, the forecast increment is
\begin{equation}
\widehat{\Delta W}_{t+k,i} =
\operatorname{clip}
\left(
\beta_{b_i}^{k}\Delta W_{t,i},
-\mathrm{cap}_{b_i},
\mathrm{cap}_{b_i}
\right),
\quad k=1,\ldots,H.
\label{eq:irfio_generalized_update}
\end{equation}
Future native-cell \wse{} is then reconstructed by cumulative summation,
\begin{equation}
\widehat{W}_{t+k,i} =
W_{t,i}
+
\sum_{j=1}^{k}
\widehat{\Delta W}_{t+j,i}.
\label{eq:lrfio_forecast_reconstruction}
\end{equation}

Thus, online inference is a structured learned-operator application: the current increment field and the learned regional parameter bank jointly determine the forecast update at each native cell.

This formulation makes the machine-learning role explicit. LRFIO learns the response-field structure, regional weight sets, and retained complexity offline, then compresses the selected surrogate into a closed-form operator for deployment. The useful predictive structure is stored in the retained response case, response-field assignments, regional inertia coefficients, increment caps, and selector metadata.

LRFIO evaluates an ordered response hierarchy:
\begin{enumerate}
\item $H_0$: persistence, with zero active future increment.
\item $H_1$: global calibrated inertia, with one learned response.
\item $H_2$: segmented response-field inertia, with multiple learned responses.
\end{enumerate}

This hierarchy answers a model-selection question: how much learned response complexity is justified once the current native-cell increment is available as a solver-conditioned state signal? For $H_0$, the retained parameters are zero-active, producing no future increment. For $H_1$, all native cells map to a single global bin, $b_i=1$, so one learned weight set $\theta_1=(\beta_1,\mathrm{cap}_1)$ is applied across the domain. For $H_2$, cells map to segmented response-field bins $b_i\in\{1,\ldots,B\}$, allowing different regions to follow different learned decay and cap constraints. The retained model may select persistence, retain one global learned response, or admit segmented response-field structure depending on validation evidence.

\subsection{Offline Learning and Response-Field Construction}
\label{subsec:irfio_offline_calibration}

Offline learning separates parameter fitting, validation-based selection, and held-out testing to prevent leakage~\cite{Hastie2009_ESL,Kaufman2012_LeakageDataMining}. The candidate evidence family is
\begin{equation}
\mathcal{H}_{\mathrm{LRFIO}}
=
\left\{
H_0,\ H_1,\ H_2^{(B_1)},\ H_2^{(B_2)},\ldots
\right\},
\label{eq:irfio_candidate_family}
\end{equation}
where $B_j$ denotes a requested segmented response-field resolution.

For segmented candidates, a training-period response score is computed for each cell. A representative score is the calibration-period \wse{} range:
\begin{equation}
s_i
=
\max_{\tau\in\mathcal{T}_{\mathrm{cal}}} W_{\tau,i}
-
\min_{\tau\in\mathcal{T}_{\mathrm{cal}}} W_{\tau,i}.
\label{eq:irfio_response_score_range}
\end{equation}
This score is a learned-response descriptor that organizes cells according to their observed solver-produced variability during the calibration interval. Cells are assigned to response bins by $b_i=q_B(s_i)$.

For each bin $b$, LRFIO calibrates an inertial coefficient $\beta_b$ and an empirical increment cap $\mathrm{cap}_b$:
\begin{equation}
\mathcal{D}_b
=
\left\{
|\Delta W_{\tau,i}|
:
i \in b,\ \tau\in\mathcal{T}_{\mathrm{cal}}
\right\},
\quad
\mathrm{cap}_b=Q_q(\mathcal{D}_b).
\label{eq:irfio_cap_quantile}
\end{equation}
The coefficient $\beta_b$ learns how the current native-cell increment persists or decays across future leads for cells assigned to response bin $b$. The cap $\mathrm{cap}_b$ learns an empirical stability constraint from calibration-period increments so that rollout increments remain within a response-consistent range.

Coefficient and cap candidates are selected from finite grids that include zero-active values. Including zero-active candidates allows requested higher-order response fields to resolve naturally into simpler base cases when localized inertia is unsupported. The learning procedure therefore tests whether segmentation or neural correction earns its cost under the validation protocol.

\subsection{Base-Case-First Learned Selector}
\label{subsec:irfio_inductive_selector}

Each candidate $h\in\mathcal{H}_{\mathrm{LRFIO}}$ is scored on validation data using the selection metric defined in Section~\ref{sec:experimental_methodology}. Let $h^\star$ be the absolute-best candidate, $h_{\mathrm{base}}$ the best base case among $H_0$ and $H_1$, and $h_{\mathrm{seg}}$ the best segmented candidate.

LRFIO first measures the regret of the best base case relative to the absolute-best candidate:
\begin{equation}
R_{\mathrm{base}}
=
\frac{
S(h_{\mathrm{base}})-S(h^\star)
}{
|S(h^\star)|
}.
\label{eq:irfio_base_regret}
\end{equation}
If $R_{\mathrm{base}}\leq\tau_{\mathrm{base}}$, where $\tau_{\mathrm{base}}=0.02$, the base case is deemed sufficient and retained. If the base case exceeds this regret tolerance, segmentation is considered. The segmented model is admitted only if its gain over the best base case reaches the structural threshold $\tau_{\mathrm{seg}}=0.05$:
\begin{equation}
G_{\mathrm{seg}}
=
\frac{
S(h_{\mathrm{base}})-S(h_{\mathrm{seg}})
}{
|S(h_{\mathrm{base}})|
}.
\label{eq:irfio_segmented_gain}
\end{equation}
The final learned selection rule is
\begin{equation}
h_{\mathrm{LRFIO}}
=
\begin{cases}
h_{\mathrm{base}},
&
R_{\mathrm{base}}\leq\tau_{\mathrm{base}}, \\[3pt]
h_{\mathrm{seg}},
&
R_{\mathrm{base}}>\tau_{\mathrm{base}}
\ \wedge
G_{\mathrm{seg}}\geq\tau_{\mathrm{seg}}, \\[3pt]
h_{\mathrm{base}},
&
\text{otherwise}.
\end{cases}
\label{eq:irfio_selector}
\end{equation}
This rule permits small controlled regret for simpler base cases and admits segmented response-field complexity only when validation evidence justifies the added structure.

The selector is part of the learning contribution. It performs validation-based complexity control over the response-operator family and selects the response case supported by the declared information-access policy. The selected response case is therefore an ablation-minimal learned operator for the dataset.

\subsection{Relation to Neural-Layer Ablation}
\label{subsec:irfio_neural_layer_ablation}

LRFIO is motivated by an ablation-driven surrogate-modeling result. A neural correction layer can be added to an increment-based inertia scaffold to learn residual updates or response adjustments through standard forward propagation. Such a layer is retained when it improves held-out accuracy or the speed--accuracy tradeoff relative to the learned closed-form response operator.

The LRFIO formulation treats the neuralized inertia variant as a candidate family in the ablation space. If the neural layer improves validation and held-out behavior under the same information-access policy, it can be reported as a stronger candidate. If the closed-form response operator preserves or improves accuracy while reducing deployment cost, the retained model is the ablation-minimal learned response operator selected by the validation protocol.

This distinction is central to the interpretation of the method. The learned quantities in LRFIO are the response structure, bin assignments, inertial coefficients, increment caps, and selector decision. The online computation is intentionally lightweight because the learning has already been compressed into the retained parameter bundle. This makes LRFIO appropriate for repeated rollout settings where deployment cost is as important as accuracy.

\subsection{Parameter Bundle and Online Deployment}
\label{subsec:irfio_deployment}

The offline stage yields a compact dataset-specific learned parameter bundle:
\begin{equation}
\Theta^{(d)}_{\mathrm{LRFIO}}
=
\left\{
h_{\mathrm{LRFIO}},
\mathbf{b},
\boldsymbol{\beta},
\mathbf{cap},
\mathcal{A}_{\mathrm{sel}}
\right\},
\label{eq:irfio_parameter_bundle}
\end{equation}
where $\mathcal{A}_{\mathrm{sel}}$ stores selector-audit metadata. The bundle stores the learned response case, cell-to-response assignments, retained inertial coefficients, retained increment caps, and the validation evidence used to justify the selected complexity.

During deployment, the selector is not re-run. LRFIO accesses only the current \wse{}, the previous \wse{}, the resulting current increment, and the learned bundle $\Theta^{(d)}_{\mathrm{LRFIO}}$. Parameters are retrieved through the lookup map
\begin{equation}
i \mapsto b_i \mapsto (\beta_{b_i},\mathrm{cap}_{b_i}).
\label{eq:irfio_lookup_structure}
\end{equation}
Future increments are generated in closed form using \eqref{eq:irfio_generalized_update}, and future native-cell \wse{} fields are reconstructed by cumulative summation. LRFIO therefore selects the retained response complexity for each dataset and deploys the selected learned operator through the same compact rollout form.

The result is a learned surrogate with asymmetric cost: offline calibration and ablation learn the response operator, while online deployment performs only the retained closed-form rollout. This separation is useful for \hecras{} emulation because it preserves the benefits of data-driven model selection while enabling sub-second forecast generation in repeated deployment scenarios.
% ============================================================

% ============================================================

\section{Experimental Methodology}
\label{sec:experimental_methodology}

\subsection{Evaluation Strategy and Candidate Families}
\label{subsec:method_cross_dataset_strategy}

The experimental methodology evaluates surrogate-model components across multiple \hecras{} 2D datasets rather than comparing model names on a single benchmark. The goal is to determine which learned modeling choices consistently support fast, solver-consistent \wse{} prediction while preserving the information-access constraints defined in Section~\ref{sec:problem_data_protocol}. The central experimental question is how much learned model capacity is justified once the current native-cell increment is retained as a solver-conditioned response signal.

The evaluated candidate space is organized around representation, forecast protocol, update form, learned response structure, and information access. Representation choices include native cells, rasters, graphs, and reduced-order summaries. Forecast protocols include closed-form, direct-horizon, and autoregressive rollout. Update forms include direct-state prediction, increment prediction, inertial continuation, residual correction, and response-field propagation. Information-access conditions distinguish no-forcing models from forcing-aware variants.

To establish how much predictive skill is available from the current state before adding model complexity, the benchmark includes a native-cell learned-response ladder:
\begin{enumerate}
\item Persistence and constant-increment continuation.
\item Fixed-beta and globally capped inertia.
\item Globally calibrated inertia.
\item Segmented response-field calibrated inertia.
\item Neuralized inertia and learned residual variants.
\end{enumerate}
This ladder supports component-level interpretation. Persistence and constant-increment continuation define simple current-state references. Global and segmented calibrated inertia test whether a compact learned response operator captures the dominant predictive structure. Neuralized inertia and residual variants test whether online learned correction improves the speed--accuracy tradeoff beyond the closed-form response operator.

These native-cell references and learned-response variants are compared against higher-capacity surrogate families, including raster, graph, recurrent, neural-operator, and forcing-conditioned models. For paper-level reporting, implementation rows are grouped into conceptual model families. The retained Learned Response-Field Inertia Operator (LRFIO) row for each dataset is the representative selected by the base-case-first learned selection rule in Section~\ref{sec:irfio}. The selected representative reflects validation-supported response complexity rather than the most complex candidate in the grid or the candidate with the largest number of trainable parameters.

\subsection{Protocol and Information-Access Metadata}
\label{subsec:method_protocol_metadata}

All models are evaluated using the canonical native-cell arrays and chronological splits defined in Section~\ref{sec:problem_data_protocol}. To prevent models with different input information from being interpreted as direct architecture-only alternatives~\cite{Kaufman2012_LeakageDataMining}, each evaluated row is registered with a compact joint forcing/protocol label:
\begin{equation}
\begin{aligned}
\mathrm{NF\mbox{-}AR} &: \text{no forcing, stepwise/autoregressive},\\
\mathrm{NF\mbox{-}DH} &: \text{no forcing, direct or closed-form},\\
\mathrm{F\mbox{-}SA} &: \text{forcing, step-ahead/autoregressive},\\
\mathrm{F\mbox{-}FH} &: \text{forcing, full-horizon/direct}.
\end{aligned}
\label{eq:method_access_labels}
\end{equation}
These labels distinguish no-forcing models from those using prescribed scenario-definition inputs~\cite{USACE_HECRAS_BoundaryConditions}. This distinction is necessary because a full-horizon-forcing neural model and a no-forcing current-state response operator solve different prediction problems even if both output future \wse{} fields.

The retained LRFIO model is evaluated strictly as an NF-DH model. It uses the current \wse{}, previous \wse{}, current increment, and the learned parameter bundle retained before test evaluation. Neuralized inertia variants are evaluated under the same information-access condition when included in the ablation study, so that any difference in accuracy or runtime reflects model structure rather than extra future forcing information.

\subsection{Ablation Design and Learned-Operator Selection}
\label{subsec:method_ablation_design}

The ablation design evaluates whether added model complexity improves held-out behavior relative to simpler learned-response operators. The ablation proceeds from simple to complex candidates. First, persistence and constant-increment references establish the baseline predictive value of the current state. Second, global and capped inertia candidates test whether a single learned response coefficient and stability cap are sufficient. Third, segmented response-field candidates test whether different native-cell response regions require different learned inertia parameters. Fourth, neuralized inertia or residual-correction candidates test whether online learned correction improves the retained response operator beyond the calibrated closed-form form.

This design establishes the methodological status of LRFIO as a learned surrogate selected through validation evidence. The validation protocol identifies the response structure that provides the best justified speed--accuracy tradeoff under the declared information-access policy. Neural correction remains part of the candidate space when it improves validation and held-out behavior. When the closed-form response operator preserves or improves accuracy while reducing deployment cost, the retained model is the ablation-minimal learned response operator.

The ablation separates three questions that are often conflated:
\begin{enumerate}
\item Does the current native-cell increment provide a strong predictive scaffold?
\item Does learned response calibration improve over fixed or uncalibrated inertial continuation?
\item Does additional segmentation or neural correction improve enough to justify its added complexity?
\end{enumerate}
LRFIO is the retained response-operator family produced by this ablation logic.

\subsection{Metrics, Selection Score, and Selector Audit}
\label{subsec:method_metrics_selection_score}

Standard error summaries such as global \rmse{} can hide final-lead, high-tail, or localized error behavior~\cite{Chai2014_RMSE_MAE}. Therefore, candidates are ranked using an aggregate validation score $\mathcal{J}$ that penalizes broad-domain accuracy, final-lead behavior, area-weighted behavior, hotspot error, upper-tail error, and bias:
\begin{equation}
\begin{aligned}
\mathcal{J}
={}&
1.10\,R_{\mathrm{stage}}
+0.10\,R_H
+0.05\,Q_{95}\!\left(|e_H|\right) \\
&\quad
+0.05\,R_{\mathrm{Top100},H}
+0.05\,|B_{\mathrm{stage}}|.
\end{aligned}
\label{eq:selection_score}
\end{equation}
where $R_{\mathrm{stage}}$ is stage \rmse{}, $R_H$ is final-lead \rmse{}, $R_{\mathrm{Top100},H}$ is top-100 final-lead hotspot \rmse{}, and $B_{\mathrm{stage}}$ is stage bias.

The weights are fixed across all datasets and candidates: the broad-domain stage \rmse{} term receives an effective weight of $1.10$, final-lead \rmse{} receives weight $0.10$, final-lead $Q_{95}(|e_H|)$ receives weight $0.05$, top-100 final-lead hotspot \rmse{} receives weight $0.05$, and absolute stage bias receives weight $0.05$. Lower values are better.

To make retained complexity explicit, a selector audit tracks the LRFIO validation process. Using the rules defined in Section~\ref{subsec:irfio_inductive_selector}, the audit records the absolute-best candidate, the best base candidate, the best segmented candidate, segmented gain over the base case, regret relative to the absolute best, and the final selection reason. This audit ensures that segmented response-field complexity is retained only when the validation-score improvement satisfies the declared structural-gain threshold.

The selector audit also clarifies the machine-learning interpretation of the method. The retained LRFIO model is a validation-selected learned operator whose structure is determined from data. The selected case may be persistence, global calibrated inertia, or segmented response-field inertia depending on the dataset. The retained model therefore learns the simplest response structure supported by the evidence for the declared information-access condition.

\subsection{Runtime and Speedup Reporting}
\label{subsec:method_runtime_reporting}

Following standard performance-evaluation practice, runtime is reported separately from training and calibration costs to answer the operational question of deployment efficiency~\cite{Jain1991_PerformanceEvaluation}. For surrogate models, runtime is measured as wall-clock seconds per forecast rollout window under the evaluated deployment protocol. For LRFIO, this includes reading current and previous \wse{}, forming the current increment, retrieving retained learned parameters, and computing the closed-form multi-lead forecast.

Training and calibration costs are treated separately because the proposed workflow has asymmetric cost: offline learning estimates and selects the response operator, while online deployment evaluates the retained operator cheaply. This distinction is especially important for the neural-layer ablation. A neuralized candidate may have additional training and inference costs, whereas the retained closed-form LRFIO compresses the learned response structure into a compact parameter bundle. Runtime reporting therefore evaluates the deployment consequence of the ablation decision.

For \hecras{} numerical solves, runtime is reported as a horizon-normalized speedup estimate when a measured solve time is available. If the measured solve time corresponds to a complete simulation, the solve time is normalized to the evaluated forecast horizon by multiplying by the ratio between forecast duration and complete simulated duration. Speed--accuracy comparisons therefore plot deployment runtime per forecast window against stage \rmse{}, with more favorable models falling toward the lower-left region.

The resulting comparison is a speed--accuracy evaluation of learned surrogate families for \hecras{} output emulation. LRFIO is evaluated as a learned no-forcing current-state emulator whose retained response operator provides competitive native-cell accuracy with low deployment cost. The runtime analysis therefore measures the operational value of compressing the learned response structure into a lightweight closed-form rollout.
% ============================================================

% ============================================================

\section{Results}
\label{sec:results}

This section reports the measured outcomes of the cross-dataset native-cell benchmark. The retained architecture is the Learned Response-Field Inertia Operator (LRFIO), evaluated as a no-forcing, closed-form native-cell rollout under the methodology defined in Section~\ref{sec:experimental_methodology}. LRFIO deploys through a closed-form update, and the retained operator is produced by offline surrogate learning, validation-based response selection, and ablation over competing response structures.

The results support four main findings:
\begin{enumerate}
\item \textbf{Learned adaptive complexity:} LRFIO selects dataset-specific response complexity. It retains segmented response-field inertia for Beaver Bayou, global calibrated inertia for Upper San Saba and Tuttle Creek, and persistence for Lower San Saba.
\item \textbf{Ablation-controlled model selection:} The selector audit demonstrates that added response-field complexity is retained only when validation evidence justifies it. The global base case is retained for Tuttle Creek with 4.30\% regret because the segmented alternative does not meet the 5\% structural-gain threshold.
\item \textbf{Fast learned-operator deployment:} Retained rollout times range from 0.003~s to 0.242~s across the datasets, with an estimated $2.75 \times 10^4$ horizon-normalized speedup over the measured \hecras{} solve on Beaver Bayou.
\item \textbf{Favorable speed--accuracy tradeoffs:} The LRFIO family occupies the most favorable runtime--accuracy region against the evaluated comparator families, supporting the interpretation that the current native-cell increment is a strong solver-conditioned scaffold for learned surrogate rollout.
\end{enumerate}

\subsection{Retained LRFIO Decisions}
\label{subsec:results_retained_lrfio_decisions}

Table~\ref{tab:results_final_lrfio_selected_rows} reports the strict retained LRFIO candidate for each dataset. The selector outcomes are dataset-dependent: Beaver Bayou retains segmented response-field inertia ($H_2$) because the segmented gain meets the retained threshold. Upper San Saba and Tuttle Creek retain global calibrated inertia ($H_1$), while Lower San Saba retains persistence ($H_0$).

\begin{table*}[!t]
\caption{Final retained LRFIO response cases and held-out test accuracy. $H_0$ denotes persistence, $H_1$ global calibrated inertia, and $H_2$ segmented response-field inertia.}
\label{tab:results_final_lrfio_selected_rows}
\centering
\renewcommand{\arraystretch}{1.12}
\footnotesize
\setlength{\tabcolsep}{5pt}
\begin{tabular*}{\textwidth}{@{\extracolsep{\fill}}lccccc@{}}
\toprule
\textbf{Dataset} &
\textbf{Case} &
\textbf{Active seg.} &
\textbf{Stage \rmse{} (m)} &
\textbf{Final \rmse{} (m)} &
\textbf{Selector outcome} \\
\midrule
Beaver Bayou &
$H_2$ segmented &
2 &
0.082421 &
0.143701 &
Seg. gain $\geq 5\%$ \\

Upper San Saba River &
$H_1$ global &
1 &
0.006830 &
0.010248 &
Base within $2\%$ \\

Lower San Saba River &
$H_0$ persistence &
0 &
0.000060 &
0.000070 &
Base within $2\%$ \\

Tuttle Creek / Big Blue / Kansas River &
$H_1$ global &
1 &
0.133756 &
0.154906 &
Seg. gain $<5\%$ \\
\bottomrule
\end{tabular*}
\end{table*}

These outcomes show that LRFIO learns and selects a response structure from training and validation evidence for each project. The retained parameter bundles are compact. Beaver Bayou retains two active response-field segments with $\beta$ values ranging from 0.82 to 0.97 and caps ranging from 0.011515~m to 0.036980~m. Upper San Saba retains one global response with $\beta=0.97$ and $cap=0.001221$~m. Lower San Saba retains zero-active persistence. Tuttle Creek retains one global response with $\beta=0.76$ and $cap=2.382294$~m. These values report the learned response structure and calibrated parameters used for closed-form rollout.

\subsection{Selector Audit and Ablation-Controlled Regret}
\label{subsec:results_selector_audit}

Table~\ref{tab:results_lrfio_selector_audit} reports the selector audit. Three of the four retained candidates are also the absolute-best internal-validation candidates. The exception, Tuttle Creek, illustrates the utility of the base-case-first learned selector. While the absolute-best candidate for Tuttle Creek was segmented inertia with \texttt{n\_bins\_12}, its gain over the best base case was approximately 4.12\%---below the retained 5\% threshold. The selector therefore retained the simpler global inertia base case, accepting a controlled regret of 4.30\% relative to the absolute-best score.

Across all four datasets, the mean regret fraction is approximately 1.07\%, and the maximum regret fraction is 4.30\%. The retained decisions match the independently audited selector outcomes exactly, confirming that the final models are produced by the declared learned-selection rule.

\begin{table*}[!t]
\caption{LRFIO selector audit. Regret is reported relative to the absolute-best internal-validation candidate. Segmented gain is measured relative to the best base case.}
\label{tab:results_lrfio_selector_audit}
\centering
\renewcommand{\arraystretch}{1.14}
\footnotesize
\setlength{\tabcolsep}{4pt}
\begin{tabular*}{\textwidth}{@{\extracolsep{\fill}}lcccccc@{}}
\toprule
\textbf{Dataset} &
\textbf{Selected} &
\textbf{Absolute best} &
\textbf{Best base} &
\textbf{Best segmented} &
\textbf{Regret} &
\textbf{Seg. gain} \\
\midrule
Beaver Bayou &
\texttt{n\_bins\_2} &
\texttt{n\_bins\_2} &
\texttt{n\_bins\_1} &
\texttt{n\_bins\_2} &
0.0000 &
0.1024 \\

Upper San Saba River &
\texttt{n\_bins\_1} &
\texttt{n\_bins\_1} &
\texttt{n\_bins\_1} &
-- &
0.0000 &
-- \\

Lower San Saba River &
\texttt{n\_bins\_0} &
\texttt{n\_bins\_0} &
\texttt{n\_bins\_0} &
-- &
0.0000 &
-- \\

Tuttle Creek / Big Blue / Kansas River &
\texttt{n\_bins\_1} &
\texttt{n\_bins\_12} &
\texttt{n\_bins\_1} &
\texttt{n\_bins\_12} &
0.0430 &
0.0412 \\
\bottomrule
\end{tabular*}
\end{table*}

This audit is an ablation result over the learned response hierarchy. It shows that LRFIO retains segmentation only when its validation gain exceeds the structural threshold; otherwise, the simpler learned response case is selected. The retained closed-form operator is therefore an ablation-minimal learned surrogate selected from the evaluated response hierarchy.

\subsection{Deployment Runtime and Speedup}
\label{subsec:results_runtime_speedup}

Table~\ref{tab:results_lrfio_runtime_speedup} reports retained LRFIO deployment runtimes, measuring closed-form rollout only and excluding offline calibration. Retained rollout times scale consistently with domain size and forecast horizon, ranging from 0.003~s for Beaver Bayou to 0.242~s for Upper San Saba.

\begin{table*}[!t]
\caption{Deployment runtime and available \hecras{} speedup evidence for retained LRFIO candidates.}
\label{tab:results_lrfio_runtime_speedup}
\centering
\renewcommand{\arraystretch}{1.14}
\footnotesize
\setlength{\tabcolsep}{4pt}
\begin{tabular*}{\textwidth}{@{\extracolsep{\fill}}lccccc@{}}
\toprule
\textbf{Dataset} &
\textbf{Response case} &
\textbf{Cells} &
\textbf{Horizon steps} &
\textbf{LRFIO runtime (s)} &
\textbf{Speedup evidence} \\
\midrule
Beaver Bayou &
Segmented inertia &
9,236 &
24 &
0.003188 &
$2.75\times10^{4}$ horizon-normalized \\

Upper San Saba River &
Global inertia &
430,874 &
48 &
0.241594 &
-- \\

Lower San Saba River &
Persistence &
201,676 &
48 &
0.064116 &
-- \\

Tuttle Creek / Big Blue / Kansas River &
Global inertia &
44,255 &
37 &
0.021221 &
-- \\
\bottomrule
\end{tabular*}
\end{table*}

A measured full \hecras{} numerical solve time is available for Beaver Bayou, which required 431~s for a 59~h simulation. Because the retained LRFIO evaluation uses a 12~h forecast horizon, the horizon-normalized \hecras{} runtime is 87.66~s. Dividing this by the retained LRFIO runtime of 0.003188~s yields a horizon-normalized speedup estimate of approximately $2.75\times10^4$. Using the full measured solve time without horizon normalization gives a raw solve-to-surrogate ratio of approximately $1.35\times10^5$; the horizon-normalized value is the more conservative comparison.

The runtime result reflects the asymmetric cost structure of the method. LRFIO performs learning, calibration, and response selection offline, then deploys the retained learned operator through a lightweight closed-form update. This is the practical consequence of the ablation-driven design: when the retained response operator provides the best justified tradeoff, the learned response structure can be compressed into a cheaper online deployment form.

\subsection{Speed--Accuracy Tradeoff Against Comparator Families}
\label{subsec:results_speed_accuracy_tradeoff}

\begin{figure*}[!t]
\centering
\includegraphics[width=\textwidth]{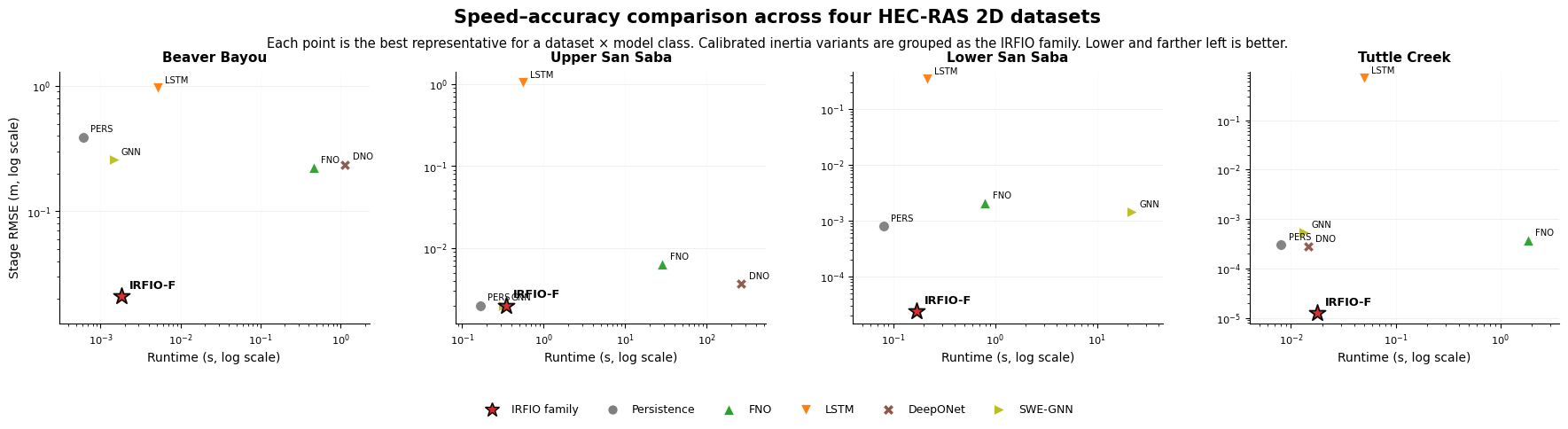}
\caption{Speed--accuracy comparison across four \hecras{} 2D datasets.
Each point is the best representative for a dataset--model-class pair under the shared held-out test protocol.
Both axes use logarithmic scaling. Lower and farther left is better.
The LRFIO-family representative occupies the most favorable speed--accuracy region across the evaluated comparator families.}
\label{fig:speed_accuracy_lrfio_family}
\end{figure*}

Figure~\ref{fig:speed_accuracy_lrfio_family} and Table~\ref{tab:results_speed_accuracy_summary} compare held-out stage \rmse{} against measured deployment runtime across the evaluated model families. Each point represents the best family-level representative for a dataset--model-class pair. The comparison is intentionally reported at the model-family level rather than as a raw implementation leaderboard; minor implementation variants, calibration-grid variants, and aliases are grouped when they express the same underlying surrogate family.

\begin{table*}[!t]
\caption{Compact numerical summary of Fig.~\ref{fig:speed_accuracy_lrfio_family}.
For each dataset, the LRFIO-family representative is compared with the strongest published non-LRFIO comparator by held-out stage \rmse{}. Lower values are better.}
\label{tab:results_speed_accuracy_summary}
\centering
\renewcommand{\arraystretch}{1.12}
\footnotesize
\setlength{\tabcolsep}{3.5pt}
\begin{tabular*}{\textwidth}{@{\extracolsep{\fill}}lccccc@{}}
\toprule
\textbf{Dataset} &
\multicolumn{2}{c}{\textbf{LRFIO family}} &
\textbf{Best published} &
\multicolumn{2}{c}{\textbf{Best published non-LRFIO}} \\
\cmidrule(lr){2-3} \cmidrule(lr){5-6}
&
\textbf{\rmse{} (m)} &
\textbf{Time (s)} &
\textbf{non-LRFIO} &
\textbf{\rmse{} (m)} &
\textbf{Time (s)} \\
\midrule
Beaver Bayou &
0.021192 &
0.001846 &
FNO &
0.222007 &
0.454737 \\

Upper San Saba &
0.001997 &
0.343371 &
SWE-GNN &
0.001997 &
0.323601 \\

Lower San Saba &
0.000024 &
0.169520 &
SWE-GNN &
0.001442 &
21.998938 \\

Tuttle Creek &
0.000013 &
0.017852 &
DeepONet &
0.000274 &
0.014644 \\
\bottomrule
\end{tabular*}
\end{table*}

Across the evaluated datasets, several published comparator families are competitive on individual cases. However, they generally occupy less favorable regions of the plot because they require more runtime, produce larger error, or both. In contrast, the LRFIO family remains consistently near the lower-left frontier across all four panels.

Table~\ref{tab:results_speed_accuracy_summary} shows that the LRFIO-family representative gives the lowest stage \rmse{} on three datasets and ties the strongest published comparator on Upper San Saba. This tie provides useful context: the comparison is best read as a speed--accuracy tradeoff rather than as a claim that LRFIO is always simultaneously fastest and most accurate. Overall, these results show that a learned response-field inertia operator is competitive with substantially more complex surrogate families while operating at low deployment cost.

\subsection{Residual and Neuralized-Inertia Ablation}
\label{subsec:results_residual_neuralized_inertia_ablation}

The previous results compare the retained LRFIO family against broader external surrogate families. This subsection isolates a narrower architectural question: whether the retained closed-form Learned Response-Field Inertia Operator should be augmented with additional online learned correction capacity. Two learned-correction variants are evaluated against the retained LRFIO rollout. The first, LRFIO-R, adds a ridge residual corrector to the retained LRFIO prediction. The second, INO, is an isolated neuralized-inertia operator that uses the same native-cell, no-forcing direct-horizon information-access condition but replaces the lightweight residual corrector with an MLP residual head over the learned inertial scaffold. This comparison is motivated by the broader use of neural operators, graph surrogates, and deep hydrodynamic surrogate models in recent flood-modeling literature~\cite{Li2021_FNO,Lu2021_DeepONet,Li2023_GeoFNO,Li2023_GINO,Pfaff2021_MeshGraphNets,Bentivoglio2023_SWEGNN,Bentivoglio2025_mSWEGNN,HacesGarcia2023_DLHFRAN,Sun2023_FNOFlood}. The residual-correction baseline also provides a regularized learned-correction test consistent with standard statistical learning practice~\cite{Hastie2009_ESL}. Accuracy is reported using held-out stage \rmse{}, and runtime is interpreted as deployment cost following standard performance-measurement practice~\cite{Chai2014_RMSE_MAE,Jain1991_PerformanceEvaluation}.

\begin{figure*}[!t]
\centering
\includegraphics[width=\textwidth]{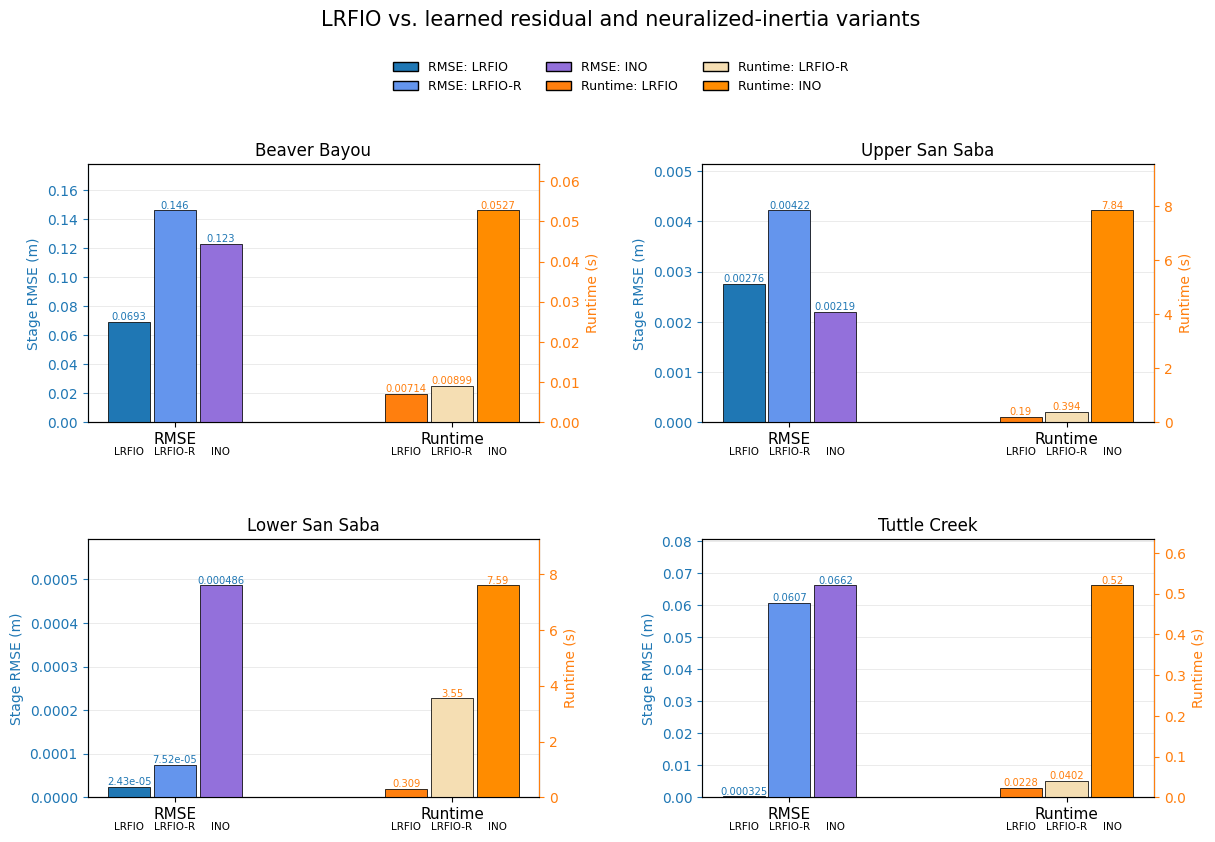}
\caption{Direct learned-correction ablation comparing the retained Learned Response-Field Inertia Operator (LRFIO), a ridge residual-correction variant (LRFIO-R), and an isolated neuralized-inertia operator (INO). Each panel reports held-out stage \rmse{} on the left axis and deployment runtime on the right axis for one dataset. All three variants are evaluated under the no-forcing direct-horizon information-access condition. Lower values are better for both metrics.}
\label{fig:lrfio_lrfior_ino_ablation}
\end{figure*}

\begin{table*}[!t]
\caption{Direct residual and neuralized-inertia ablation. Ratios are computed relative to retained LRFIO. Values greater than one indicate that the learned-correction variant is worse than LRFIO for the corresponding metric.}
\label{tab:lrfio_lrfior_ino_ablation}
\centering
\renewcommand{\arraystretch}{1.12}
\footnotesize
\setlength{\tabcolsep}{3.2pt}
\begin{tabular*}{\textwidth}{@{\extracolsep{\fill}}lcccccc@{}}
\toprule
\textbf{Dataset} &
\textbf{LRFIO} &
\textbf{LRFIO-R} &
\textbf{INO} &
\textbf{LRFIO} &
\textbf{LRFIO-R} &
\textbf{INO} \\
&
\textbf{\rmse{} (m)} &
\textbf{\rmse{} ratio} &
\textbf{\rmse{} ratio} &
\textbf{time (s)} &
\textbf{time ratio} &
\textbf{time ratio} \\
\midrule
Beaver Bayou &
0.069288 &
2.108 &
1.777 &
0.007143 &
1.258 &
7.379 \\

Upper San Saba River &
0.002761 &
1.528 &
0.794 &
0.189558 &
2.079 &
41.355 \\

Lower San Saba River &
0.000024 &
3.102 &
20.025 &
0.309338 &
11.468 &
24.543 \\

Tuttle Creek / Big Blue / Kansas River &
0.000325 &
186.919 &
203.905 &
0.022833 &
1.759 &
22.791 \\
\bottomrule
\end{tabular*}
\end{table*}

Figure~\ref{fig:lrfio_lrfior_ino_ablation} and Table~\ref{tab:lrfio_lrfior_ino_ablation} show that the retained closed-form LRFIO provides the strongest overall speed--accuracy tradeoff. LRFIO-R increases held-out stage \rmse{} on all four datasets and also increases runtime on all four datasets. The lightweight ridge residual correction therefore weakens the retained tradeoff. The isolated neuralized-inertia operator is more nuanced: it improves stage \rmse{} on Upper San Saba, reducing the LRFIO-relative \rmse{} ratio to 0.794, but this improvement requires a runtime increase of approximately $41.36\times$. On the other three datasets, INO is both less accurate and slower than LRFIO, with especially large \rmse{} penalties on Lower San Saba and Tuttle Creek.

This ablation supports the retained architecture by showing that the evaluated online learned-correction variants do not provide a consistently favorable tradeoff under the same native-cell support, chronological split, and no-forcing direct-horizon information-access condition. The retained LRFIO is therefore an ablation-minimal learned operator: the response structure, inertial coefficients, increment caps, and retained complexity are learned offline, while deployment is compressed into a closed-form native-cell update. In this benchmark, adding a ridge residual corrector or an MLP neuralized-inertia residual head adds online cost without improving the retained cross-dataset tradeoff.

\subsection{Interpretation as an Ablation-Minimal Learned Operator}
\label{subsec:results_ablation_minimal_operator}

Taken together, the selector audit, retained parameter bundles, and speed--accuracy comparison support the central ablation claim. The dominant predictive structure in these native-cell \hecras{} tasks is captured by the current solver-produced \wse{} increment and a learned response rule governing how that increment persists, decays, saturates, or varies across response regions. Additional complexity is useful only when it improves the retained validation tradeoff.

The retained LRFIO model is an ablation-minimal learned surrogate operator. Learning occurs during offline calibration and validation-based selection, where the method estimates the response structure, inertial coefficients, increment caps, and retained complexity. The online stage is intentionally inexpensive because the learned surrogate has been compressed into the selected response operator.

% ============================================================

\section{Discussion}
\label{sec:discussion}

The cross-dataset results support an ablation-driven interpretation of native-cell \hecras{} 2D surrogate modeling. By preserving the current native-cell \wse{} increment as a solver-conditioned response signal, LRFIO learns an inertial response operator that operates near the favorable runtime--accuracy frontier through compact closed-form deployment. The retained model is selected from a response hierarchy rather than imposed as a fixed segmented model or deep spatial architecture. This makes the final operator a validation-supported response structure whose complexity is determined by the evidence available for each dataset.

This interpretation is central to the method. LRFIO learns its retained response case, cell-to-response assignments, inertial coefficients, increment caps, and selection decision from solved \hecras{} trajectories before held-out testing. The resulting online update is closed form because the learned response structure has been compressed into a compact parameter bundle. For the evaluated native-cell \wse{} prediction task, the calibrated response operator captures the dominant predictive structure efficiently enough that additional online neural correction is not retained by the ablation protocol.

\subsection{Interpretation of the Retained Learned Response and Increment Scaffold}
\label{subsec:discussion_retained_response_cases}

The dataset-specific retained cases in Table~\ref{tab:results_final_lrfio_selected_rows} validate the base-case-first learned selector rule. Lower San Saba retains persistence, Upper San Saba and Tuttle Creek retain global calibrated inertia, and Beaver Bayou retains segmented response-field inertia. This variation shows that the retained response structure is selected from data for each project. Instead of adopting a lowest-validation-error rule that can favor added structure even for marginal gains~\cite{Hastie2009_ESL}, LRFIO requires added complexity to satisfy a declared structural-gain threshold. The Tuttle Creek audit illustrates this point: the segmented candidate has the lowest internal-validation score, but its gain over the best base case remains below the retained threshold, so the selector retains global calibrated inertia.

The success of these compact learned-response models stems from the solver-consistent setting. The current native-cell increment, $\Delta \mathbf{W}_t$, is the immediate response of the \hecras{} model after numerical routing through terrain, roughness, storage, and boundary constraints~\cite{USACE_HECRAS_NumericalMethods}. Spatial heterogeneity is already encoded in this increment. LRFIO's learned parameters, $\beta$ and $cap$, approximate how the solver-produced increment persists, decays, or saturates over future leads. This is conceptually related to local-inertial shallow-water approximations, while remaining a learned surrogate update rather than a flux-solving hydraulic scheme~\cite{Bates2010_SimpleInertial,Almeida2013_LocalInertialApplicability}.

The key methodological implication is that the current solver-produced increment can serve as a strong feature representation for native-cell surrogate learning. When this response signal is available at forecast initialization, model capacity can be directed toward learning how the increment should be propagated, constrained, or segmented. LRFIO exploits this structure by learning the response rule that governs future rollout from the current native-cell increment.

\subsection{Relationship to Neural, Higher-Capacity, and Forcing-Aware Models}
\label{subsec:discussion_higher_capacity_forcing}

The evaluated candidate space includes neural-operator, graph, recurrent, raster, and reduced-order surrogate families motivated by established approaches~\cite{Li2021_FNO,Lu2021_DeepONet,Pfaff2021_MeshGraphNets,Bentivoglio2023_SWEGNN,Bentivoglio2025_mSWEGNN,Hochreiter1997_LSTM,Cho2014_GRU,Ronneberger2015_UNet,Shi2015_ConvLSTM,Schmid2010_DMD,Dutta2021_PODRBF_NIROM}. The results place these higher-capacity methods in a controlled comparison against strong native-cell learned inertial baselines. This comparison is important because persistence alone is a weak reference for evaluating whether added spatial, graph, recurrent, or operator capacity is justified.

The neural-layer ablation should be interpreted in this same component-evaluation framework. A neural correction layer is useful when it improves the retained accuracy--runtime tradeoff under the same information-access condition. In the evaluated benchmark, the retained closed-form operator provides the stronger cross-dataset tradeoff, so the ablation protocol selects the learned response operator compressed into closed-form deployment. This outcome preserves the machine-learning role of the method: learning occurs in the response structure, parameter calibration, and validation-based selection, while online inference applies the retained learned operator efficiently.

Architectural comparisons must also account for representation burden, deployment complexity, and information access. This distinction is critical for forcing-aware models. Boundary hydrographs and lateral inflows are standard scenario-definition inputs~\cite{USACE_HECRAS_BoundaryConditions}. A model conditioned on full-horizon future forcing solves a different deployment problem than a current-state no-forcing model. Apparent gains in forcing-aware models may therefore reflect additional scenario information, architecture, or both~\cite{Kaufman2012_LeakageDataMining}. Consequently, LRFIO's reported performance should be interpreted specifically as no-forcing current-state emulation.

\subsection{Implications for Native-Cell HEC-RAS Surrogate Learning}
\label{subsec:discussion_implications}

Methodologically, this study suggests that native-cell \hecras{} 2D surrogate modeling should begin by testing how much future solver response is already encoded in the current state. A practical evaluation hierarchy follows:
\begin{enumerate}
\item Establish persistence and constant-increment references.
\item Test fixed-beta and capped-inertia scaffolds.
\item Test global calibrated inertia.
\item Admit segmented response-field inertia when justified by structural gain.
\item Add neural correction, graph, raster, Fourier, recurrent, or modal complexity when it demonstrably improves the accuracy--runtime tradeoff beyond these strong learned-response baselines.
\end{enumerate}

This hierarchy is a disciplined evaluation procedure rather than a theoretical error floor. LRFIO may not yield the lowest possible \rmse{} for every future scenario, but it provides a strong learned-response baseline for determining whether additional complexity is necessary. The retained sub-second rollout times and the available Beaver Bayou speedup evidence indicate that LRFIO is suitable for repeated emulation tasks such as rapid scenario screening, ensemble triage, and sensitivity analysis~\cite{Razavi2012_SurrogateReview}. It should be interpreted as an efficient learned emulator of \hecras{} outputs within the declared no-forcing current-state protocol.

The broader implication is that surrogate-modeling studies should report whether simpler learned operators have already captured the dominant predictive signal before attributing gains to higher-capacity architectures. For this benchmark, the evidence indicates that the native-cell increment response is a strong predictive scaffold. The main contribution is therefore an ablation-supported finding about where the useful predictive structure resides: in the solver-conditioned current increment and the learned response rule that propagates it.

\subsection{Limitations and Future Work}
\label{subsec:discussion_limitations_future_work}

Several limitations define the boundaries of this benchmark. First, cross-dataset evaluation demonstrates architectural consistency across four \hecras{} projects, while LRFIO parameters remain dataset-specific in the present study. Second, the target is solver-consistent \wse{} emulation rather than field-truth hydraulic validation. Third, target variables such as velocity, face flux, momentum, sediment, breach processes, and strong wet/dry transitions remain outside the current scope~\cite{USACE_HECRAS_2DUsersManual,USACE_HECRAS_NumericalMethods,USACE_HECRAS_BoundaryConditions}. Finally, reported deployment runtimes exclude offline calibration, which still requires solved \hecras{} trajectories~\cite{Jain1991_PerformanceEvaluation}.

Future work should extend this benchmark to multi-event settings involving hydrograph magnitude and timing shifts, geometry variants, and structure operations. A forcing-aware LRFIO extension is a natural progression, provided it remains explicitly labeled to preserve the information-access distinction. A promising direction is to strengthen native-cell response diagnostics, using peak timing, recession slope, wetness history, cell area, roughness, local slope, or connectivity descriptors to define event-conditioned parameter bundles.

Another important direction is cross-project transfer. The present article learns a dataset-specific response operator for each project. A stronger generalization study would use leave-one-project-out evaluation, training response-selection policies or hyperparameter priors on three \hecras{} projects and testing adaptation to a fourth. Such an experiment would determine whether LRFIO can move beyond per-dataset calibration toward transferable learned response-operator families.

Finally, future work should extend the learned-response framework beyond the present dataset-specific calibration setting. The benchmark already includes forcing-aware, feature-rich, graph, raster, recurrent, neural-operator, residual, and neuralized-inertia alternatives, and the retained LRFIO speed--accuracy tradeoff remains competitive against these additional inputs and model classes. This outcome motivates richer conditioning strategies that improve the learned response operator itself. Future work should therefore test whether event class, geometry, forcing context, graph neighborhoods, or longer temporal histories can condition $\theta_b=(\beta_b,\mathrm{cap}_b)$ in a transferable way across events or projects while preserving native-cell support, chronological splits, and explicit information-access labels.

% ============================================================

% ============================================================

\section{Conclusion}
\label{sec:conclusion}

This article presented a cross-dataset evaluation of learned native-cell surrogate models for solver-consistent water-surface elevation (\wse{}) prediction in \hecras{} 2D unsteady-flow simulations. By enforcing an explicit information-access policy~\cite{Kaufman2012_LeakageDataMining,USACE_HECRAS_WorkingWithProjects,USACE_HECRAS_BoundaryConditions}, the study separates architectural effects from scenario-forcing access across four benchmark datasets. This framing is essential for surrogate learning because no-forcing current-state emulation, forcing-aware prediction, and full-horizon neural prediction define distinct deployment conditions.

The primary contribution is the Learned Response-Field Inertia Operator (LRFIO), a no-forcing, increment-based learned surrogate. LRFIO learns an inertial response operator from solved \hecras{} trajectories and uses a base-case-first selector to evaluate an ordered hierarchy of response hypotheses: persistence, global calibrated inertia, and segmented response-field inertia. The retained response case adapts by dataset, selecting segmented response-field inertia for Beaver Bayou, global calibrated inertia for Upper San Saba and Tuttle Creek, and persistence for Lower San Saba. The selector audit further shows that added response-field complexity is retained only when validation evidence satisfies the declared structural-gain threshold.

The ablation result is that effective surrogate learning for these native-cell \hecras{} tasks can be compressed into a compact learned response operator. LRFIO learns response structure, inertial coefficients, increment caps, and selector decisions offline, then deploys the retained operator through a closed-form rollout. The final model is therefore an ablation-minimal learned operator whose useful predictive structure has been compressed into an inexpensive deployment rule.

The results support LRFIO as a fast native-cell emulation method. Retained rollout times remain below one second across all evaluated datasets, and the Beaver Bayou comparison gives an estimated $(2.75\times10^4)$ horizon-normalized speedup over the measured \hecras{} solve. The speed--accuracy comparison further shows that the LRFIO family remains near the favorable runtime--accuracy frontier against the evaluated comparator families.

The methodological implication is that the current native-cell increment is a strong solver-conditioned predictive scaffold~\cite{USACE_HECRAS_NumericalMethods}. This increment reflects recent solver response through terrain, storage, roughness, boundary conditions, and numerical controls. Learned inertial-response baselines should therefore be established before attributing predictive gains to higher-capacity spatial architectures. Added neural, graph, raster, modal, or segmented complexity should be retained when it improves the accuracy--runtime tradeoff under the same information-access condition.

The conclusions remain bounded by the study scope. The surrogate target is solver-consistent \hecras{} output emulation, and the evaluated target variable is \wse{}. Velocity, face flux, sediment, hydraulic-structure operation, uncertainty, and strong wetting/drying transitions remain outside the current benchmark scope~\cite{USACE_HECRAS_2DUsersManual,USACE_HECRAS_NumericalMethods,USACE_HECRAS_BoundaryConditions}. The retained LRFIO parameters are also dataset-specific in the present study. Future work should extend the framework to multi-event scenarios, explicitly labeled forcing-aware variants, stronger native-cell response diagnostics, neuralized inertia ablations, and leave-one-project-out transfer experiments while preserving the central discipline that added modeling complexity must earn its cost.

% ------------------------------------------------------------
% Bibliography
% ------------------------------------------------------------
\begingroup
\sloppy
\bibliographystyle{IEEEtran}
\bibliography{references}
\endgroup

% ============================================================
\end{document}